# On the Semantics and Automated Deduction for PLFC, a Logic of Possibilistic Uncertainty and Fuzziness

**Teresa Alsinet**
Computer Science Dept.
Universitat de Lleida (UdL)
25080 Lleida, Spain
tracy@eup.udl.es

**Lluís Godo**
AI Research Inst. (IIIA)
CSIC
08193 Bellaterra, Spain
godo@iiia.csic.es

**Sandra Sandri**
CS and Applied Math. Lab. (LAC)
INPE
S. José dos Campos 12201-970, Brazil
sandri@lac.inpe.br

**Abstract**

inconsistent Recently, a syntactical extension of first order Possibilistic logic (called PLFC) dealing with fuzzy constants and fuzzily restricted quantifiers has been proposed. In this paper we present steps towards both the formalization of PLFC itself and an automated deduction system for it by (i) providing a formal semantics; (ii) defining a sound resolution-style calculus by refutation; and (iii) describing a first-order proof procedure for PLFC clauses based on (ii) and on a novel notion of most general substitution of two literals in a resolution step.

## 1 INTRODUCTION

The necessity-valued fragment of Possibilistic logic [Dubois et al., 1994] is a logic of uncertainty to reason under incomplete information and partially inconsistent knowledge, built upon classical first order logic. There exists for PL a proof procedure based on a refutation complete resolution-style calculus. Recently, in [Dubois et al., 1998] a syntax of an extension of PL (called PLFC) dealing with fuzzy constants and fuzzily restricted quantifiers has been proposed. In this paper we present steps towards both the formalization of PLFC itself and an automated deduction system for it by (i) providing a formal semantics; (ii) defining a sound resolution-style calculus by refutation – completeness issues for PLFC are out of scope; and (iii) describing a first-order proof procedure for PLFC clauses based on (ii) and on a novel notion of most general substitution of two literals in a resolution step. In contrast to standard PL semantics, truth-evaluation of formulas with fuzzy constants are many-valued instead of boolean, and consequently an extended notion of possibilistic uncertainty is also needed. Remark, however, that PLFC is an uncertainty logic and clearly departs from truth-functional fuzzy logic systems and the like [Mukaidono et al., 1989].

The paper is organized as follows. Next section is a refresh of standard Possibilistic logic, while in Section 3 we recall the extension PLFC as proposed in [Dubois et al., 1998]. Sections 4 and 5 deal with all the necessary semantical aspects of PLFC to be able to provide sound resolution rule refutation mechanisms in Section 6. Finally, based on them, in Section 7 we describe an automated deduction method for PLFC. Proofs of propositions and theorems are to be found in [Alsinet et al., 1999].

## 2 NECESSITY-VALUED POSSIBILISTIC LOGIC

In necessity-valued Possibilistic logic each formula is represented by a pair $(\varphi, \alpha)$, $\varphi$ being a classical, closed first order logic formula and $\alpha \in (0, 1]$ being a lower bound on the belief on $p$ in terms of necessity measures. A formula $(\varphi, \alpha)$ is thus interpreted as a constraint $N(\varphi) \geq \alpha$, where $N$ is a necessity measure on propositions, a mapping from the set of logical formulae to a totally ordered bounded scale, usually (but not necessarily) given by $[0, 1]$, characterized by the axioms

(i)     $N(\top) = 1$,
(ii)    $N(\bot) = 0$,
(iii)   $N(\varphi \wedge \psi) = min(N(\varphi), N(\psi))$,
(iv)   $N(\varphi) = N(\psi)$,
        if $\varphi$ and $\psi$ are classically equivalent,

where $\top$ and $\bot$ denote respectively tautology and contradiction.

The necessity-valued Possibilistic logic (simply Possibilistic logic from now on) is axiomatized (Hilbert-style) by the axioms of classical f.o. logic weighted by 1, together with the following graded versions of the usual *modus ponens* and *generalization* inference rules,

$$\frac{(\varphi,\alpha), (\varphi \to \psi, \beta)}{(\psi, \min(\alpha,\beta))}[MP], \quad \frac{(\varphi,\alpha)}{((\forall x)\varphi,\alpha)}[G]$$

together with a *weight weakening* rule

$$\frac{(\varphi,\alpha)}{(\varphi,\beta)}[W]$$

for $\beta \leq \alpha$. We shall denote by $\vdash_{PL}$ the notion of proof in Possibilistic logic derived from this formal system of axioms and rules.

Resolution by refutation is an automated deduction method which has been nicely adapted to Possibilistic logic. Indeed, let $K$ be a knowledge base formed by possibilistic clauses, i.e. possibilistic formulas of the type $(\psi,\alpha)$, where $\psi$ is a (f.o. or propositional) clause in the usual sense. Then, it holds that a formula $(\varphi,\alpha)$ is derived in Possibilistic logic from the knowledge base $K$, i.e. $K \vdash_{PL} (\varphi,\alpha)$, iff we obtain a proof of $(\bot,\alpha)$ by successively applying the below *resolution* rule in $K \cup \{(\neg \delta_i, 1) \mid i=1,n\}$, where $\bigvee_{i=1,n} \delta_i$ is the clausal form of $\varphi$:

$$\frac{(\neg p \vee q, \alpha), (p \vee r, \beta)}{(q \vee r, \min(\alpha,\beta))}[Res]$$

In other words, if we denote the above procedure of proof by refutation through resolution by $\vdash^r_{PL}$, we have that $K \vdash_{PL} (\varphi,\alpha)$ iff $K \vdash^r_{PL} (\varphi,\alpha)$. Moreover, using the $\vdash^r_{PL}$ procedure, other rules can be derived, for instance the *fusion* rule

$$\frac{(p,\alpha), (p,\beta)}{(p, max(\alpha,\beta))}.$$

Now, let us recall here the usual (monotonic) semantics for possibilistic logic. For the sake of an easier understanding we first consider the propositional case, and after the first order case.

**The propositional case.** *Notation*: Let $L$ be a propositional language and let $\Omega$ be the set of classical interpretations for $L$, that is, the set of evaluations $w$ of the atoms of the language into the boolean truth value set $\{0,1\}$. Each evaluation of atoms $w$ extends to any clause in the usual way, and thus for any $\varphi$, $w(\varphi) \in \{0,1\}$. For any clause $\varphi$, we will write $w \models \varphi$ iff $w(\varphi) = 1$. We shall also write $[\varphi]$ to denote the set of models of $\varphi$, i.e. $[\varphi] = \{w \in \Omega \mid w \models \varphi\}$.

Belief states are modelled by normalized possibility distributions $\pi : \Omega \to [0,1]$ on the set of possible interpretations. A possibility distribution $\pi$ is normalized when there is at least one $w \in \Omega$ such that $\pi(w) = 1$. In other words, belief states modelled by normalized distributions are consistent states, in the sense that at least one interpretation (or state or possible world) has to be fully plausible. These will be our possibilistic models. The satisfaction relation between possibilistic models (i.e. possibility distributions) and possibilistic formulas is defined as follows:

$$\pi \models (\varphi,\alpha) \text{ iff } N([\varphi] \mid \pi) \geq \alpha,$$

where $N(. \mid \pi)$ is the necessity measure induced by $\pi$ on the power set of $\Omega$, defined as $N([\varphi] \mid \pi) = \inf_{w \in \Omega} \max(1 - \pi(w), w(\varphi)) = \inf_{w \not\models \varphi} 1 - \pi(w)$. If $\pi \models (\varphi,\alpha)$ we say that $\pi$ is a model of $(\varphi,\alpha)$. An interesting equivalent expression is:

$$\pi \models (\varphi,\alpha) \text{ iff for all } w \in \Omega, \pi(w) \leq \max(1-\alpha, w(\varphi)),$$

As usual, if $\Gamma$ denotes a set of possibilistic clauses, we say that $\pi$ is a model of $\Gamma$ iff $\pi$ is a model of each formula in $\Gamma$. The possibilistic entailment, denoted $\models_{PL}$, is then defined as follows.

$$\Gamma \models_{PL} (\varphi,\alpha) \text{ iff } \pi \models (\varphi,\alpha),$$

for each $\pi$ being model of $\Gamma$. [Dubois et al., 1994] shows that this semantics makes Possibilistic logic sound and complete, and moreover, using refutation, the resolution-based proof system briefly sketched in the previous subsection is also sound and complete wrt to the above semantics, that is,

$$\Gamma \models_{PL} (\varphi,\alpha) \text{ iff } \Gamma \vdash_{PL} (\varphi,\alpha) \text{ iff } \Gamma \vdash^r_{PL} (\varphi,\alpha)$$

**First order case.** When the language $L$ is of first order, things do not change very much. Possibilistic f.o. formulas are of the type $((\forall x \ldots \forall y)\varphi(x \ldots y), \alpha)$, where $\varphi(x \ldots y)$ is a clause with free variables $x \ldots y$. First order interpretations are structures $w = (U, i, m)$, where U is a domain (or sets of domains if we have sorts), $i$ maps each predicate of arity $n$ to a subset of $U^n$ and $m$ maps each object constant to an element of the domain $U$. Then, if $\varphi$ is a closed f.o. formula, we continue writing $w \models \varphi$ to denote that $\varphi$ is true in the interpretation $w$. Now, possibilistic models are possibility distributions $\pi$ on the set of f.o. interpretations $\Omega$ of $L$, and possibilistic satisfaction and entailment are then just as in the propositional case. Completeness results for first order possibilistic logic are also provided in [Dubois et al., 1994].

## 3 PL + FUZZY CONSTANTS + VARIABLE WEIGHTS

In [Dubois et al., 1998] the authors propose an extension of possibilistic logic, always using formulas in clausal form, where, in order to deal with fuzzy predicates and ill-known values, variable weights and generalized fuzzy constants are allowed respectively. In the rest of this paper we shall refer to this extension as PLFC.

Variable weights are employed to enable the modelling of statements such as "the more $x$ is $A$ (or $x$ belongs to $A$), the more certain is $p(x)$", where $A$ is a fuzzy set. This is formalized as, for all $x$, "$p(x)$ is true with a necessity of at least $\mu_A(x)$", and represented as

$$(p(x), A(x))$$

where it is assumed, as in PL, that all variables appearing in a formula are universally quantified. When $A$ is imprecise but not fuzzy, the interpretation of such a formula is just "$\forall x \in A, p(x)$". So $A$ acts as a (flexible if it is fuzzy) restriction on the universal quantifier.

Fuzzy constants are used to model typical fuzzy statements of the type "in Brazil the mean temperature in December is *about_25*", represented as

$$mean\_temp(Brazil, december, about\_25),$$

where *mean_temp* is a classical predicate and *about_25* is a generalized constant. If *about_25* denotes a crisp interval of temperatures, then the above expression is interpreted as $\exists x \in about\_25$ such that $mean\_temp(Brazil, december, x)$ is true. So, fuzzy constants can be seen as (flexible) restrictions on an existential quantifier. In general, "$L(B)$ is true at least to degree $\alpha$", will be represented as $(L(B), \alpha)$, where $L$ is either a positive or negative literal and $B$ is a fuzzy set.

Special attention must therefore be given when reading a literal such as $\neg p(B)$ in a possibilistic formula. For instance, if $B$ is not fuzzy, $(p(B), 1)$ and $(\neg p(B), 1)$ have to read as "$\exists x \in B, p(x)$" and "$\exists x \in B, \neg p(x)$" respectively. Moreover, a formula like $(\neg p(B) \vee r(A), 1)$, where $A$ and $B$ are not fuzzy, should be interpreted as "it is certain that $\exists x \in B, \exists y \in A, p(x) \rightarrow r(y)$", that is completely different of "it is certain that $\forall x \in B, \exists y \in A, p(x) \rightarrow r(y)$", which would be represented as $(\neg p(x) \vee r(A), B(x))$[1].

In order to deal with variable weights and fuzzy constants, the following inference pattern was used

$$\frac{(\neg p(x) \vee q(x), A(x)), (p(B) \vee r, \alpha)}{(q(B) \vee r, \min(\alpha, N(A \mid B)))}[RR]$$

where $N(A \mid B) = \inf_{x \in X} \max(1 - \mu_B(x), \mu_A(x))$ is a necessity-like measure of how much certain is $A$ given $B$. Details on how this pattern was justified can be found in [Dubois et al., 1998]. Moreover, during the resolution process, variables may disappear in the logical part of a clause, but still appear in its valuation side. The following pattern was proposed to deal with this situation:

$$\frac{(p(y), f(x, y))}{(p(y), \max_{x \in X} f(x, y))}[FR]$$

where $f(x, y)$ is a valid valuation in the model, involving variables $x$ and $y$.

The underlying idea in [Dubois et al., 1998] was to propose an extension of Possibilistic logic sticking to classical logic proof procedures as much as possible, in particular to refutation by resolution, as in standard Possibilistic logic. However there was no evaluation there about whether such a proof by refutation method can be supported by a well-defined semantics.

## 4 EXTENDING PL SEMANTICS

We are concerned in providing PLFC with a sound semantics, extending the one provided for PL. So the matter is what has to be modified in the standrad PL semantics to support the extension of the logical constructs of PLFC. For example, consider the previously mentioned predicate

$$mean\_temp(Brazil, december, about\_25).$$

Our intended interpretation is that *about_25* is a fuzzy set describing temperatures around $25^oC$, in the range from $-50^oC$ to $50^oC$, and with a particular membership function $\mu_{about\_25} : [-50, 50] \rightarrow [0, 1]$. In doing so, as far as we can see, we are introducing two major changes in the standard semantics of PL: (i) the truth evaluations of predicates and formulas are no longer boolean but many-valued, the set of truth-values becomes the whole unit interval $[0, 1]$; and (ii) the certainty evaluation of formulas in a possibilistic model has to be extended in a suitable manner, that is, we have to define what does $N(\varphi \mid \pi)$ mean when $\varphi$ contain fuzzy constants.

**(i) From boolean to many-valued.** With respect to standard Possibilistic logic, the main difference of truth evaluations is that now, for instance, in a particular interpretation, the predicate $mean\_temp(Brazil, december, about\_25)$ can be true (1), false (0), but also can take some intermediate truth degree, depending on how much the actual temperature in the interpretation fits with the fuzzy set *about_25*. For instance, consider the interpretation $w_1 = (U, i_1, m_1)$, where $U = U_1 \times U_2 \times U_3$, $U_1$ is a set of countries, $U_2$ is the set of months and $U_3 = [-50, 50]$,

$i_1(mean\_temp) = \{$(Brazil, january, 30), ...,
 (Brazil, december, 27), ..., (Spain, january, 5),
 ..., (Spain, december, 10) $\}$,

---
[1]Notice that an expression like "it is certain that $\forall x \in B, \exists y \in A, p(x) \wedge r(y)$" cannot be represented in PLFC since it would require the use of Skolem functions, which have not yet been treated in this framework.

and where $m$ not only assigns elements of $U$ to usual object constants, but also a fuzzy set (i.e. a membership function) to each fuzzy constant, in this case to *about_25*. Then it seems natural to take as truth degree of the formula *mean_temp(Brazil, december, about_25)*, in the interpretation $w_1$, the value

$$w_1(mean\_temp(Brazil, december, about\_25)) = \\ = \mu_{m(about\_25)}(27) \quad (= 0.7, \text{ say}).$$

**(ii) Certainty-evaluation.** According to (i), we assume from now on that the truth evaluation of PLFC formulas $\varphi$ in any interpretation $w$ is a value $w(\varphi) \in [0, 1]$. Therefore each PLFC formula does not induce anymore a crisp set of interpretations, but a fuzzy set of interpretations $[\varphi]$, defining $\mu_{[\varphi]}(w) = w(\varphi)$, for any $w$. Hence, if we want to continue measuring the uncertainty induced on a PLFC formula by a possibility distribution on the set of interpretations, we have to consider some extension for fuzzy sets (of interpretations) of the standard notion of necessity measure. The basic question is, given a belief state modelled by a possibility distribution $\pi$, how to establish the possibilistic semantics of statements of the type

$$A \text{ is } \alpha - \text{certain}$$

where $A$ is a fuzzy set. We want to define a measure $N(. \mid \pi)$, extension of the one previously introduced for classical sets, in such a way that a possibility distribution $\pi$ supports the statement iff $N(A \mid \pi) \geq \alpha$. This question has already been tackled by Dubois and Prade (see f.i. [Dubois et al., 1994]) where they propose to use this index:

$$N^*(A \mid \pi) = \inf_{w \in \Omega} \pi(w) \Rightarrow \mu_A(w)$$

where $\Rightarrow$ is the reciprocal of Gödel many-valued implication, defined as $x \Rightarrow y = 1$, if $x \leq y$, $x \Rightarrow y = 1 - x$, otherwise. But the bad news about this candidate semantics is that proof by refutation (using the resolution rule of the previous section) is not sound, even though the resolution rule itself can be proved to be sound. Let us consider the following PLFC clauses

$$A_i = mean\_temp(Brazil, december, \mu_i),$$

$i = 1, 2$, where $\mu_1$ and $\mu_2$ are trapezoidal fuzzy sets[2] of temperatures defined as $\mu_1 = [20; 24; 26; 30]$ and $\mu_2 = [20; 25; 25; 30]$. It is easy to check that $\inf_{t \in [-50, 50]} \mu_1(t) \Rightarrow \mu_2(t) = 0$, thus $(A_2, \alpha)$ cannot be a logical consequence of $(A_1, 1)$ if $\alpha > 0$. On the other hand, by refutation, using the resolution

---
[2]We use the representation of a trapezoidal fuzzy set as $[t_1; t_2; t_3; t_4]$ where the interval $[t_1, t_4]$ is the support and the interval $[t_2, t_3]$ is the core.

---

rule RR introduced in the previous section, we get that $\{(A_1, 1), (\neg A_2(x), \mu_2(x))\} \vdash (\bot, \beta)$, where $A_2(x)$ stands for $mean\_temp(Brazil, december, x)$, and $\beta = N(\mu_2 \mid \mu_1) = \inf_{t \in [-50, 50]} \max(1 - \mu_1(t), \mu_2(t)) = 4/9 > 0$.

However, there is an alternative notion of necessity of fuzzy event which is commonly used in Possibility Theory as a measure of pattern matching [Dubois and Prade, 1988], which is to define

$$N(A \mid \pi) = \inf_{w \in \Omega} \max(1 - \pi(w), \mu_A(w)).$$

This definition also extends the standard notion of necessity degree when $A$ is crisp, and we have $N(A \mid \pi) = 1$ only when every plausible interpretation ($\pi(w) > 0$) makes $A$ totally true ($\mu_{A(w)} = 1$). Now, the condition $N(A \mid \pi) \geq \alpha$ becomes equivalent to the inequality

$$\pi(w) \leq \max(1 - \alpha, \mu_{[A]_\alpha}(w))$$

for all $w \in \Omega$, where $[A]_\alpha$ denotes the $\alpha$-cut of $A$, i.e. $[A]_\alpha = \{w \in \Omega \mid \mu_A(w) \geq \alpha\}$. We will show later that using this semantics a sound refutation-based proof procedure can be defined. Moreover, there is a nice axiomatization for the above defined necessity measure for fuzzy sets. Namely, let $\Omega$ be a set and let $N : [0, 1]^\Omega \to [0, 1]$ be a measure on the set of fuzzy sets of $\Omega$. Consider the following postulates:

**N1**    $N(\Omega) = 1$  
**N2**    $N(\emptyset) = 0$  
**N3**    $N(A \cap B) = \min(N(A), N(B))$  
       ( $N(\cap_{i \in I} A_i) = \inf_{i \in I} N(A_i)$ )  
**N4**    $N(A \cup \alpha) = \max(\alpha, N(A))$, if $A$ is crisp

where $\mu_{A \cap B}(w) = \min(\mu_A(w), \mu_B(w))$ for all $w \in \Omega$, and $A \cup \alpha$ denotes the fuzzy set defined by the membership function $\mu_{A \cup \alpha}(w) = \max(\alpha, \mu_A(w))$, for all $w \in \Omega$.

**Theorem 1** *If $N$ satisfies the above postulates, then there exists $\pi : \Omega \to [0, 1]$ such that, for all fuzzy subset $A$ of $\Omega$, $N(A) = \inf_{w \in \Omega} \max(1 - \pi(w), \mu_A(w))$.*

## 5    FORMALIZING PLFC

As already pointed out, in general our PLFC formulas will be pairs of the form $(\varphi(\bar{x}), f(\bar{y}))$, where $\bar{x}$ (respec. $\bar{y}$) denotes a set of variables $x_i$ (respec. $y_j$), for instance, $(p(A, x) \vee q(y), \min(\alpha, \mu_B(y), \mu_C(y)))$. The left-hand side of the pair $\varphi(\bar{x})$ is a disjunction of literals, possibly with free variables $\bar{x}$ and possibly with fuzzy constants. The right-hand side $f(\bar{y})$, $\bar{y} \supseteq \bar{x}$, consists of a valuation function, defined for a superset of the free variables in the left-hand side, denoting a (variable) lower bound for the necessity value of the formula of the left-hand side.

In the next subsection we describe the language and semantics of the formulas appearing in the left-hand side. We shall refer to them as *base formulas*, and their language, *base language*, denoted as PLFC*. This is needed in the rest of the section where we define a possibilistic semantics of PLFC clauses, first with constant weight, and later with variable weight.

## 5.1 THE BASE LANGUAGE PLFC*: Semantics

The basic components of the language PLFC* are:
1. *sorts* of variables (we will distinguish a basic sort $\sigma$ from its corresponding (fuzzy) extended sort $f\sigma$, see below); a *type* is a tuple of sorts;
2. a set $X$ of object *variables*, a set $C$ of *object constants*, and a set $FC$ of *fuzzy constants*, each having its sort; furthermore a set $FC_{cuts}$ of imprecise constants corresponding to the $\alpha$-cuts of the fuzzy constants of $FC$: if $A \in FC$, then $[A]_\alpha \in FC_{cuts}$, for any $0 \leq \alpha \leq 1$.
3. a set *Pred* of *regular predicates*, each one having a type;
4. *connectives* $\neg$ and $\vee$.

*Atomic formulas* have the form $p(x,...,y)$ where $p$ is a predicate from *Pred*, $x,...,y$ are variables, object or fuzzy constants and the sorts of $x,...,y$ correspond to the type of $p$. *Literals* are of the form $p(x,...,y)$ or $\neg p(x,...,y)$, where $p$ is a predicate from *Pred*. Finally, *Clauses* are disjunctions of literals, either positive or negative. Free variables in a clause are implicitly universally quantified. For the sake of simplicity, functions are not dealt in the present framework.

Next we define the semantics of PLFC*, which, due to the presence of fuzzy constants, is many-valued, instead of boolean (two-valued).

**Definition 1 (Interpretations)** *An* interpretation $w = (U, i, m)$ *maps:*
1. *each basic sort $\sigma$ into a non-empty domain $U_\sigma$ and each extended sort $f\sigma$ into the set $F(U_\sigma)$ of fuzzy sets of $U_\sigma$;*
2. *a predicate $p$ of type $(\sigma_1, \ldots, \sigma_k, f\sigma_{k+1}, \ldots, f\sigma_n)$ into a* crisp *relation $i(p) \subseteq U_{\sigma_1} \times \ldots \times U_{\sigma_n}$;*
3. *an object constant $c$ of sort $\sigma$ into a value $m(c) \in U_\sigma$, a fuzzy constant $A$ of sort $f\sigma$ into a (normalized) fuzzy set of $m(A) \in F(U_\sigma)$, and a cut $[A]_\alpha$ into the $\alpha$-cut of $m(A)$. We will denote by $\mu_{m(.)}$ the membership function of $m(.)$. The value $m(c)$ is also represented by fuzzy set, given by $\mu_{m(c)}(m(c)) = 1$, $\mu_{m(c)}(u) = 0$, $\forall u \in U, u \neq m(c)$.*

**Definition 2 (Truth evaluation)** *An* evaluation *of variables is a mapping $e$ associating an element $e(x) \in U_\sigma$ to each variable $x$ of sort $\sigma$ or $f\sigma$. We define by cases the* truth value *of a clause under an interpretation $w = (U, i, m)$ and an evaluation of variables $e$:*

1. $w_e(p(\ldots, x, \ldots, c)) =$
$\sup_{(\ldots, u, \ldots, v, \ldots) \in i(p)} \min(\ldots, \mu_{e(x)}(u), \ldots, \mu_{m(c)}(v), \ldots)$
2. $w_e(\neg p(\ldots, x, \ldots, c, \ldots)) =$
$\sup_{(\ldots, u, \ldots, v, \ldots) \notin i(p)} \min(\ldots, \mu_{e(x)}(u), \ldots, \mu_{m(c)}(v), \ldots)$
3. $w_e(\varphi \vee \psi) = \max(w_e(\varphi), w_e(\psi))$

*Finally, the truth-value of a clause under an interpretation $w$, is defined as $w(\varphi) = \inf_e w_e(\varphi)$.*

It is clear that $w_e(\varphi)$ may take any intermediate value between 0 and 1 as soon as $\varphi$ contains some fuzzy constant. Moreover, notice that the negation in this semantics is not truth-functional.

**Example 1** Let $price(.,.)$ be a binary predicate of type $(\sigma_1, f\sigma_2)$, let $prod_1, prod_2, prod_3$ be constants of type $\sigma_1$ and let $about_{35}$ be a constant of type $f\sigma_2$. Further, let $w_0 = (U, i_0, m)$ be an interpretation such that:
1. $U = (U_{\sigma_1} = \{potatoes, apples, salad\},$
$U_{\sigma_2} = [0, 100](ptas.));$
2. $i_0(price) =$
$\{(potatoes, 40), (apples, 50), (salad, 25)\};$
3. $m(prod_1) = potatoes, m(prod_2) = apples,$
$m(prod_3) = salad, m(x) = x,$ for any $x \in [0, 100],$
$m(about\_35) = [25; 35; 35; 45];$

Now consider the ground atom $price(prod_1, about\_35)$. The truth evaluations in $w_0$ of the corresponding positive and negative literals are:

$w_0(price(prod_1, about\_35)) =$
$= \sup_{(u,v) \in i_0(price)} \min(\mu_{m(prod_1)}(u), \mu_{m(about\_35)}(v))$
$= \min(\mu_{m(prod_1)}(potatoes), \mu_{[25;35;35;45]}(40)) = 0.5$

$w_0(\neg price(prod_1, about\_35)) =$
$= \sup_{(u,v) \notin i_0(price)} \min(\mu_{m(prod_1)}(u), \mu_{m(about\_35)}(v))$
$= \sup_{x \neq 40} \min(1, \mu_{m(about\_35)}(x)) = 1$ □

## 5.2 POSSIBILISTIC SEMANTICS FOR PLFC: Clauses with Constant Weights

Let us consider PLFC clauses of the form $(\varphi, \alpha)$, where $\alpha \in [0, 1]$. Assume that $\varphi$ contains fuzzy constants, for instance take $\varphi$ to be $p(A)$, where $A$ is a fuzzy constant. In order to measure the certainty of $p(A)$ in a possibilistic model, we need to assume that $A$ has a fixed interpretation, in terms of its membership function, otherwise we would not be able to compute its necessity degree. Therefore, when defining the possibilistic models as possibility distributions over interpretations, we cannot take into account any possible interpretation, but only those which share a common and particular interpretation of the fuzzy constants, and hence they also have to share the domain. This leads us to the following notion of *context*.

**Definition 3 (Context)** *Let $U$ be a domain and let $m$ be an interpretation of object constants over $U$ (or over $[0,1]^U$ in the case of fuzzy constants). We further assume that $U$ and $m$ are such that $m$ interprets each object (non-fuzzy) constant into a (different) element of the domain $U$. We define the* context *determined by $U$ and $m$, denoted $\Omega_{U,m}$, as the set*

$$\Omega_{U,m} = \{w \in \Omega \mid w = (U, i, m)\}.$$

Now we are ready to introduce the notion of possibilistic satisfaction in a context.

**Definition 4 (Possibilistic satisfaction)** *Given a context $\Omega_{U,m}$, and a possibility distribution $\pi : \Omega_{U,m} \to [0,1]$, we define:*

$$\pi \models_{PLFC}^{U,m} (\varphi, \alpha) \text{ iff } N([\varphi] \mid \pi) \geq \alpha,$$

*where the extended necessity measure is defined as*

$$N([\varphi] \mid \pi) = \inf_{w \in \Omega_{U,m}} \max(1 - \pi(w), w(\varphi)).$$

When clear from the context, the subscript and superscript of $\models_{PLFC}^{U,m}$ will be dropped.

**Example 2** (Example 1 contd.) Consider the context $\Omega_{U,m}$ where $U$ and $m$ are as in Example 1, consider the following predicate interpretation mappings

$i_0(price) = \{(potatoes, 40), (apples, 50), (salad, 25)\}$
$i_1(price) = \{(potatoes, 35), (apples, 45), (salad, 20)\}$
$i_2(price) = \{(potatoes, 45), (apples, 50), (salad, 30)\}$

and denote $w_0 = (U, i_0, m), w_1 = (U, i_1, m), w_2 = (U, i_2, m)$. Finally, consider the following possibility distribution $\pi$ on the context $\Omega_{U,m}$: $\pi(w_0) = 0.6$, $\pi(w_1) = 1$, $\pi(w_2) = 0.2$, and $\pi(w) = 0$ otherwise. Let us now compute how much $\pi$ makes $price(prod_1, about\_35)$ certain. Remember that $m(about\_35) = [25; 35; 35; 45]$, and thus we have:

$w_0(price(prod_1, about\_35)) = \mu_{m(about\_35)}(40) = 0.5$
$w_1(price(prod_1, about\_35)) = \mu_{m(about\_35)}(35) = 1$
$w_2(price(prod_1, about\_35)) = \mu_{m(about\_35)}(45) = 0$

Then $N(price(prod_1, about\_35) \mid \pi) =$
$= \inf_{w \in \Omega_{U,m}} \max(1 - \pi(w), w(price(prod_1, about\_35)))$
$= \min(0.5, 1, 0.8, 1) = 0.5$.

Therefore, for instance, in the current context,
$$\pi \models (price(prod_1, about\_35), 0.5)$$
but
$$\pi \not\models (price(prod_1, about\_35), 0.6). \qquad \Box$$

Some very interesting and remarkable consequences of the above definition of possibilistic satisfaction are the following ones.

**Proposition 1** *In any context $\Omega_{U,m}$, under general continuity conditions[3] of the interpretation of fuzzy constants, it holds that*

*(i)* $\pi \models (p(A), \alpha)$ *iff* $\pi \models (p([A]_\alpha), \alpha)$.
*(ii)* $\pi \models (p(A, B), \alpha)$ *iff* $\pi \models (p([A]_\alpha, [B]_\alpha), \alpha)$.
*(iii)* $\pi \models (p(A) \vee q(B), \alpha)$ *iff* $\pi \models (p([A]_\alpha) \vee q([B]_\alpha), \alpha)$.

*where $p$ and $q$ can be positive or negative literals, and $[A]_\alpha, [B]_\alpha$ denote the imprecise constants corresponding to the $\alpha$-cuts of the fuzzy constants $A$ and $B$.*

These properties have important consequences since it means that in PLFC with (only) fuzzy constants we can in a way forget about fuzzy constants as such and focus only on imprecise but crisp constants.

### 5.3 POSSIBILISTIC SEMANTICS FOR PLFC: Clauses with Variable Weights

PLFC clauses with variable weight (see Section 3) are of the general form

$$(\varphi(\bar{x}), f(\bar{y})),$$

where $\varphi(\bar{x})$ is a PLFC* clause with free variables $\bar{x}$ and $f(\bar{y})$ denotes a function with values on $[0,1]$ depending on a set of variables $\bar{y}$, $\bar{y} \supseteq \bar{x}$, in the sense that it becomes computable in a given context as soon as the variables $\bar{y}$ are instantiated. As an example, consider the general PLFC clause

$$(p(A, x) \vee q(y), \min(\alpha, B(x), C(y)),$$

where $A$, $B$ and $C$ are fuzzy constants. But to be meaningful, first of all, such a PLFC clause has to be understood under a particular context $(U, m)$, where $m$ provides meaning to the fuzzy constants by assigning them membership functions $\mu_{m(A)}, \mu_{m(B)}, \mu_{m(C)}$. Second, as already noted, free variables are assumed to be universally quantified. Thus, we shall understand the above PLFC clause as the collection of instantiated clauses with constant weight

$$\{ (p(A, c) \vee q(d), \min(\alpha, B(c), C(d))) \}_{c,d}$$

where $c, d$ vary on the set of object constants (of the corresponding type), inducing on possibilistic models $\pi$ the set of constraints[4]:

$$N(p(A, c) \vee q(d) \mid \pi) \geq \min(\alpha, B(c), C(d))$$

---

[3] It refers to the assumption that $m$ interprets fuzzy constants into left continuous membership functions. This is the case, for instance when using trapezoidal membership functions and their cuts.

[4] For the sake of a simpler notation, and since a context $\Omega_{U,m}$ will always be assumed, we will simply write, for instance, $B(c), C(d)$ instead of $\mu_{m(B)}(m(c)), \mu_{m(C)}(m(d))$.

Therefore, valuation functions $f(\bar{y})$ will be basically either constant values, or membership functions of fuzzy sets, or max-min combinations of them, or necessity measures on them. We will refer to them as *valid valuation* functions.

**Definition 5** *Given a context $\Omega_{U,m}$, and a possibility distribution $\pi : \Omega_{U,m} \to [0,1]$, we define, for each universally quantified clause $(\varphi(x), A(x))$,*

$$\pi \models_{PLFC}^{U,m} (\varphi(x), A(x)) \text{ iff,}$$
*for each object constant $c$, $\pi \models_{PLFC}^{U,m} (\varphi(c), A(c))$.*

Standard possibilistic clauses $(\varphi(x), \alpha)$, i.e. with a constant weight, can be considered as a special kind of variable weight clauses if we establish the convention of considering $\alpha$ as a fuzzy set with constant membership function $\mu_{m(\alpha)}(u) = \alpha$, for all $u \in U$.

**Proposition 2** *In any context $\Omega_{U,m}$, under general continuity conditions of the interpretation of fuzzy constants, it holds that:*

(i) *The clause $(p(A, x), \min(\alpha, B(x)))$ is semantically equivalent to $(p([A]_{\min(\alpha,B(x))}, x), \min(\alpha, B(x)))$.*

(ii) *The clause $(p(A) \vee r(x), \min(\alpha, B(x)))$ is semantically equivalent to $(p([A]_{\min(\alpha,B(x))}) \vee r(x), \min(\alpha, B(x)))$.*

(iii) *If $\pi \models (p([A]_\alpha, x), \min(\alpha, B(x)))$ then $\pi \models (p(A, x), \min(\alpha, B(x)))$.*

## 6 RESOLUTION AND REFUTATION IN PLFC

The notion of possibilistic entailment in PLFC is defined as in PL. Let $K$ be a set of possibilistic clauses $K = \{(\psi_i, f_i) \mid i = 1, n\}$. We say that $K$ entails a possibilistic clause $(\varphi, f)$ in a context $\Omega_{U,m}$, written

$$K \models_{PLFC}^{\Omega_{U,m}} (\varphi, f),$$

if, for each possibility distribution $\pi$ on $\Omega_{U,m}$, $\pi \models (\varphi, f)$ whenever $\pi \models (\psi_i, f_i)$, for all $i = 1, n$.

From now on, we shall assume a particular context $\Omega_{U,m}$ to be given, and thus, the notion of soundness will be relative to the context. Furthermore, we shall assume that $\Omega_{U,m}$ provides interpretations of fuzzy constants fulfilling the previously mentioned general continuity conditions.

The fusion rule $FR$ stated in Section 3 is obviously sound with respect of this notion of entailment, however this is not the case of the resolution-like rule $RR$. Instead, it can be shown that in PLFC the following *General resolution rule* holds:

$$\frac{(\neg p(x,b,y) \vee \psi(x,y), \min(A(x), B(y), \beta))\quad (p(C,z,r) \vee \varphi(z,r), \min(D(z), E(r), \alpha))}{(\psi(C,r) \vee \varphi(b,r), \delta(r))} [GR]$$

where $\delta(r) = \min(\beta, \alpha, B(r), E(r), D(b), N(A|[C]_{\min(D(b), E(r), \alpha)}))$.

**Theorem 2 (Soundness of GR)** *The general resolution rule GR is sound with respect to the above defined possibilistic entailment.*

It is easy to see that GR rule recovers PL resolution rule $Res$ when fuzzy constants and variable weights are not present. A couple of particular interesting cases of the GR rule are:

$$\frac{(\neg p(x) \vee \psi(x), \min(A(x), \beta)), (p(B), \alpha)}{(\psi(B), \min(N(A \mid [B]_\alpha), \beta, \alpha))} [GR_1]$$

$$\frac{(\neg p(x) \vee \psi, \min(A(x), \beta))\quad (p(y) \vee \varphi(z), \min(C(y), D(z), \alpha))}{(\psi \vee \varphi(z), \min(A(x), C(x), D(z), \beta, \alpha))} [GR_2]$$

Notice that rule $GR_1$ is the analog of the resolution rule $RR$ of Section 3, but differs from it in the term $N(A \mid [B]_\alpha)$, which in $RR$ was $N(A \mid B)$. On the other hand, if we apply the fusion rule $FR$ to the resolvent of $GR_2$, what we get is just $(\psi \vee \varphi(z), \min(Pos(A \mid C), D(z), \beta, \alpha))$, where $Pos(A \mid C) = \sup_x \min(A(x), C(x))$.

One of the main advantages of the present semantics for PLFC is that provides a sound refutation mechanism based on the following properties.

**Theorem 3 (Refutation)**

(i) $K \cup \{(\neg p(x), A(x))\} \models (\bot, \alpha)$ iff $K \models (p(A), \alpha)$.

(ii) *If $A$ is an imprecise non-fuzzy constant:*
$K \cup \{(\neg p(A), 1)\} \models (\bot, \beta)$ iff
$K \models (p(x), \min(\beta, A(x)))$.

(iii) *If $K \cup \{(\neg p(A_{>0}), 1)\} \models (\bot, \beta)$ then*
$K \models (p(x), \min(\beta, A(x)))$,
*where $A_{>0}$ denotes the support of $A$, i.e.*
$\mu_{A_{>0}}(c) = 1$ *if* $\mu_A(c) > 0$, $= 0$ *otherwise.*

(iv) *If $K \cup \{(\neg p(x, B_{>0}), A(x))\} \models (\bot, \alpha)$ then*
$K \models (p(A, x), \min(\alpha, B(x)))$.

(v) *If $K \cup \{(\neg p(x), A(x)), (\neg q(B_{>0}), 1)\} \models (\bot, \beta)$ then*
$K \models (p(A) \vee q(x), \min(\beta, B(x)))$.

These soundness results will allow us to propose in the next section a refutation-based proof procedure checking whether a knowledge base $K$ of PLFC clauses entails a given clause $(\varphi, f)$. Since the complete specification of the required *negation* of a general query clause would be cumbersome, we just describe below four particular but illustrative cases. Namely, given a

threshold $\beta$, if we want to check whether the PLFC clauses:

(i) $(p(A), \beta)$,
(ii) $(p(x), \min(\beta, B(x)))$,
(iii) $(p(A, x), \min(\beta, B(x)))$,
(iv) $(\neg p(A, x) \vee q(y, B), \min(\beta, C(x), D(y)))$

are derivable from a knowledge base $K$, we have to add, respectively, to $K$ the following PLFC clauses

¬(i) $(\neg p(x), A(x))$
¬(ii) $(\neg p(B_{>0}), 1)$
¬(iii) $(\neg p(x, B_{>0}), A(x))$
¬(iv) $\{ (p(x, C_{>0}), A(x)),\ (\neg q(D_{>0}, x), B(x)) \}$

Now, Theorem 3 guarantees that if the augmented $K$ with ¬(i) (resp. with ¬(ii), ¬(iii), ¬(iv)) derives $(\bot, \beta)$, then (i) (resp. (ii), (iii), (iv)) is a logical consequence of $K$. However, nothing is said about the converse.

# 7 AUTOMATED DEDUCTION

In this section we define an automated deduction method for PLFC based on refutation through the resolution rule GR of last section, the fusion rule FR already proposed in [Dubois et al., 1998] and introduced in Section 3, and a new generalized merging rule described bellow. We then need an algorithm that let us know when two literals $p$ and $\neg p$ can be resolved. Moreover, we need an algorithm that automatically computes the set of substitutions that must be applied on the resolvent clause. But in this framework we cannot borrow the unification concept used in classical logic. Let us consider one illustrative example. For instance, from

s1: $(\neg p(A) \vee \psi, 1)$   and   s2: $(p(A), 1)$,

which, if $A$ is not fuzzy, are interpreted resp. as

"$(\exists x \in A, \neg p(x)) \vee \psi$",   $\exists x \in A,\ p(x)$",

we can infer $\psi$ iff $A$ is a precise constant. Then, the substitution algorithm for $\neg p(A)$ and $p(A)$ must fail unless $A$ is a precise constant, even though, obviously, $p(A)\theta = p(A)\theta$ for any substitution $\theta$. Therefore, from now on, we will refer ourselves to the *most general substitution of two literals in a resolution step* (or *mgs* for short). The first part of this section is about the formalization and computation of the mgs and the second describes the proof procedure by refutation.

## 7.1 MOST GENERAL SUBSTITUTION

In the following we formally define the mgs of two literals, we describe how it must be applied to a PLFC clause and we give an algorithm for its automatic computation.

**Definition 6 (substitution)** *A substitution term of a variable is either a variable, a precise constant or an imprecise non-fuzzy constant[5]. A substitution is a mapping from variables to substitution terms, and is written as $\theta = \{x_1/t_1, ..., x_n/t_n\}$, where the variables $x_1, ..., x_n$ are different and $x_i \not\equiv t_i$, for $i = 1, n$.*

Substitutions operate on expressions. By an expression we mean a term or a possibilistic clause with a constant or a variable weight. For an expression $E$ and a substitution $\theta$, $E\theta$ stands for the result of applying $\theta$ to $E$ which is obtained by *simultaneously* replacing each occurrence in $E$ of a variable from the domain of $\theta$ by the corresponding substitution term. After applying a substitution to a possibilistic clause we can obtain in the valuation side expressions like $f_1(B)$ or $f_2(B_1, \ldots, B_n)$ being $f_1$ and $f_2$ valid valuation functions in the model and $B, B_1, \ldots, B_n$ imprecise non-fuzzy constants. Then, $f_1(B)$ is computed as $N(f_1 \mid B) = \inf_x \max(1 - \mu_B(x), f_1(x))$ and $f_2(B_1, \ldots, B_n)$ as $N(f_2 \mid \min(B_1, \ldots, B_n)) = \inf_{x_1 \ldots x_n} \max(1 - \mu_{B_1}(x_1), \ldots, 1 - \mu_{B_n}(x_n), f_2(x_1, \ldots, x_n))$.

Substitutions can be composed. Given two substitutions $\theta = \{x_1/t_1, ..., x_n/t_n\}$, $\eta = \{y_1/s_1, ..., y_m/s_m\}$ their *composition* $\theta\eta$ is defined by removing from the set $\{x_1/t_1\eta, ..., x_n/t_n\eta, y_1/s_1, ..., y_m/s_m\}$ those pairs $x_i/t_i\eta$ for which $x_i \equiv t_i\eta$ and those pairs $y_i/s_i$ for which $y_i \in \{x_1, ..., x_n\}$. We say that a substitution $\theta$ is *more general* than a substitution $\eta$ if for some substitution $\gamma$ we have $\eta = \theta\gamma$.

The next algorithm takes two PLFC literals and produces their most general substitution if they can be resolved, otherwise reports an error message. We will follow the presentation of Apt [Apt, 1990], based upon Herbrand's original algorithm, first presented by Martelli and Montanari [Martelli and Montanari, 1982], which deals with solutions of finite sets of term equations.

**Algorithm 1 (most general substitution)**

*Input:* Two literals with predicate symbol $p$ of arity $n$ of the form $\neg p(s_1, ..., s_n)$ and $p(t_1, ..., t_n)$, respectively, and such that they do not have any variable in common.

*Output:* The mgs $\theta$ if they can be resolved, and otherwise, an error message.

*Initialization:* From the pair of literals $\neg p(s_1, ..., s_n)$ and $p(t_1, ..., t_n)$ we construct a set of substitutions $S$ of the form $\{s_1/t_1, ..., s_n/t_n\}$.

---

[5]Notice that fuzzy constants are not substitution terms. This is due to Proposition 1 and 2 which safely allow us to focus only on crisp constants.

*Method:* Choose any substitution $\{s_i/t_i\}$ and perform the associated action until either $S$ remains unchanged or the algorithm fails:

1. If $s_i$ and $t_i$ are object constants, then
   - if $\forall x \in \Omega$ $\mu_{s_i}(x) = \mu_{t_i}(x)$ and $s_i, t_i$ are precise constants then, delete the substitution $\{s_i/t_i\}$ from $S$;
   - otherwise, fail.
2. If $s_i$ is an object constant and $t_i$ is a variable, then replace $\{s_i/t_i\}$ by $\{t_i/s_i\}$ in $S$.
3. If $s_i$ is a variable, then
   - if $s_i \equiv t_i$ then delete the substitution $\{s_i/t_i\}$ from $S$;
   - else, if $s_i$ has another occurrence in $S$, then
     - if $s_i$ appears in $t_i$, then fail;
     - otherwise, perform the substitution $\{s_i/t_i\}$ in every other term in $S$.

*Final treatment:* If $\{s'_1/t'_1, ..., s'_k/t'_k\}$ is the resulting set of substitutions, then $\theta := \{s'_1/t'_1, ..., s'_k/t'_k\}$. □

**Example 3** Consider the following PLFC clauses:

s1: $(\neg p(x,b) \vee \psi(x), \min(A(x), \beta))$
s2: $(p(C,y), \min(D(y), \alpha))$

By Proposition 2, s2 is equivalent to:

s2': $(p([C]_{f_1(y)}, y), f_1(y))$

where $f_1(y) = \min(D(y), \alpha)$. Hence, the output of the algorithm for $\neg p(x,b)$ and $p([C]_{f_1(y)}, y)$ is the mgs $\theta = \{x/[C]_{f_1(b)}, y/b\}$, and, finally, applying $\theta$ to "pre-resolvent" of s1 and s2,

s3: $(\psi(x), \min(A(x), \beta, D(y), \alpha))$,

we get

s3': $(\psi([C]_{f_1(b)}), \min(A([C]_{f_1(b)}), \beta, D(b), \alpha))$,

where, $A([C]_{f_1(b)}) = N(A \mid [C]_{f_1(b)})$ and $D(b) = N(D \mid b) = \mu_D(b)$. Note that, by Proposition 2, s3' is equivalent to $(\psi(C), \min(A([C]_{f_1(b)}), \beta, D(b), \alpha))$, which is exactly the resolvent of s1 and s2 when applying the GR rule. □

Before developing the refutation procedure, we stress that in PLFC an extension of the PL fusion rule (see Section 2) holds. Let us first introduce two new definitions about substitutions.

**Definition 7 (variant)** *A substitution of the form $\theta = \{x_1/t_1, ..., x_n/t_n\}$ is called a* renaming *if $t_1, ..., t_n$ are different variables. Let $E_1 = (\varphi(\overline{x}), f_1(\overline{x}))$ and $E_2 = (\varphi(\overline{y}), f_2(\overline{y}))$ be two PLFC formulas. We say that $E_1$ is a* variant *of $E_2$ iff exists a renaming $\theta$ such that no variable of $E_1$ omitted in the domain of $\theta$ appears in the range of $\theta$ and $\varphi(\overline{x})\theta \equiv \varphi(\overline{y})$.*

The following *Generalized Merging Rule* is an extension of the one proposed in [Sandri and Godo, 1999]

and it corresponds to the following pattern:

$$\frac{(\varphi(\overline{x}), f_1(\overline{x})), (\varphi(\overline{y}), f_2(\overline{y}))}{(\varphi(\overline{y}), \max(f_1(\overline{x})\theta, f_2(\overline{y})))} [GM]$$

where $\theta$ is a renaming such that $(\varphi(\overline{x}), f_1(\overline{x}))$ is a variant of $(\varphi(\overline{y}), f_2(\overline{y}))$, and $f_1(\overline{x})$ and $f_2(\overline{y})$ are valid valuations functions. It can be proved that this rule is sound, but we can show more.

**Proposition 3**
*If $\theta$ is a renaming such that $(\varphi(\overline{x}), f_1(\overline{x}))$ is a variant of $(\varphi(\overline{y}), f_2(\overline{y}))$, then the set of PLFC clauses $\{(\varphi(\overline{x}), f_1(\overline{x})), (\varphi(\overline{y}), f_2(\overline{y}))\}$ is semantically equivalent to the PLFC clause $(\varphi(\overline{y}), \max(f_1(\overline{x})\theta, f_2(\overline{y})))$.*

The usefulness of this rule can be verified by means of a simple example. Let $A = \{1,2\}, B = \{2,3\}$ and $C = \{1,2,3\}$. Let us suppose we have

s1: $(p(x), A(x))$, s2: $(p(y), B(y))$, s3: $(\neg p(C), 1)$.

Resolving s1 and s3 yields $(\bot, 0)$, which is the same result we obtain when we resolve s2 and s3. Therefore, without the use of the merging rule, we only obtain $(\bot, 0)$ as final result. However, s1 is a variant of s2 and, thus, if s1 and s2 are fused together we obtain $(p(y), f(y))$, with $f(y) = \max(A(y), B(y))$, which resolved with s3 finally yields $(\bot, f([C]_1))$, where $f([C]_1) = N(\max(A,B) \mid [C]_1) = \inf_y \max(1 - \mu_{[C]_1}(y), \mu_A(y), \mu_B(y)) = 1$.

## 7.2 REFUTATION PROCEDURE

The proof procedure implements a proof by refutation through the resolution rule GR, by constructing the most general substitution of two literals in every resolution step, applying, if necessary, the fusion rule FR to the resolvent clause and, finally, merging variant clauses through the merging rule MR. Let $K$ be a knowledge base formed by PLFC clauses, $\varphi$ a PLFC clause and $\alpha$ a threshold. In order to verify if $K$ entails $\varphi$ with a necessity of at least $\alpha$, we apply the proof procedure described next:

**function** RefutationProcedure*(K: SetPLFCclauses,*
  *$\varphi$: PLFCclause, $\alpha$: NecessityDegree)* : **boolean**
**var** *RL: SetPLFCliterals /\* set of resolved literals \*/*
  $K := K \cup$ Negation*($\varphi$)*;
  $K :=$ Fusion*(K)*;
  $K :=$ Threshold*(K,$\alpha$)*;
  $K :=$ Merging*(K)*;
  $K :=$ Equivalent*(K)*;
  $RL := \emptyset$;
  **return**(ProofProcedure*(K, $\alpha$, RL)*)
**end** RefutationProcedure

where function *Negation* returns the set of clauses obtained by negating the clause $\varphi$ as stated in Theo-rem 6; function *Fusion* applies the fusion rule FR to those clauses of $K$ such that some variable appear in the valuation side but not in the base formula; function *Threshold* eliminates from $K$ the clauses such that the valuation function cannot be evaluated to a value $\beta \geq \alpha$; function *Merging* applies the generalized merging rule GM over the knowledge base; and finally, function *Equivalent* transforms, following Proposition 1 and 2, all fuzzy constants present in base formulas of $K$ into imprecise non-fuzzy constants.

During the refutation process the GM rule must be applied after every resolution step (see example in last subsection). Therefore, the proof procedure cannot be oriented to a resolvent clause and thus the search space consists of all possible orderings of the literals in the knowledge base. Moreover, for every resolution step the proof procedure is based on chronological backtracking and the search strategy is depth-first.

**function** ProofProcedure *(K: SetPLFCclauses,*
    $\alpha$*: NecessityDegree, RL: SetPLFCliterals)*: **boolean**
**variables**
  $C_1, C_2, C$: *PLFCclause*
  $L_1, L_2$: *PLFCliteral*
  $\theta$: *mgs*
  $RL'$: *SetPLFCliterals*
  $K'$: *SetPLFCclauses*
**end variables**
**if** $((\bot, \beta) \in K$ and $\beta \geq \alpha)$ **then return***(true)*
**else**
  **for** *each clause* $C_1 \in K$ **do**
   **for** *each literal* $L_1 \in C_1$ *such that* $L_1 \notin RL$ **do**
/* *assume that $C_1$ has the general form $(\varphi \vee L_1, f_1)$, where $f_1$ is a valid valuation in the model* */
    **for** *each clause* $C_2 \in K$ **do**
     **for** *each literal* $L_2 \in C_2$ **do**
/* *assume that $C_2$ has the general form $(\psi \vee L_2, f_2)$, where $f_2$ is a valid valuation in the model* */
      **if** $(\theta = mgs(L_1, L_2)$ and $\theta \neq fail)$ **then**
       **if** $(\min(f_1, f_2)\theta \geq \alpha)$ **then**
/* *even some of the variables in $f_1$ or $f_2$ have not been instantiated, the expression $\min(f_1, f_2)\theta \geq \alpha$ fails as soon as $f_1$ or $f_2$ cannot be evaluated to a value $\beta \geq \alpha$* */
        $C := Fusion\_r(((\varphi \vee \psi)\theta, \min(f_1, f_2)\theta))$;
        $K' := \text{Merging}(K \cup \{C\})$;
        $RL' := RL \cup \{L_1\}$;
        **if** (ProofProcedure$(K', \alpha, RL'))$ **then**
         **return***(true)*;
       **end if**
  **end for** $C_1$
  **return***(false)*;
**end else**
**end** ProofProcedure

where the function *Fusion_r* applies, if necessary, the FR rule to the resolvent clause of $C_1$ and $C_2$.

**Proposition 4** *Given a particular context $\Omega_{U,m}$, the notion of proof in PLFC by refutation using the GR, FR and GM rules, written $\vdash^r_{PLFC}$, is sound wrt the PLFC semantics, that is, if $\Gamma \vdash^r_{PLFC} (\varphi, \alpha)$ then $\Gamma \models_{PLFC} (\varphi, \alpha)$.*

## 8   FUTURE WORK

Future work will be addressed in three main directions. First, the extension of the base language to allow computable functions, needed f.i. in PLFC for modeling temporal resoning [Sandri and Godo, 1999]. Second, we aim at extending the current inference system (which is obviously not complete) that would allow us to have a sound and complete entailment in PLFC. Actually, the current refutation mechanism for clauses with variable weight does not allow for completeness. And the last one concerns the time complexity of the proof procedure. As we have stated, the algorithm explores all possible resolvents from the knowledge base. So, it would be interesting to find some resolution refinement in order to reduce the branching factor for the search tree. In particular, it would be nice to check for subsumed literals in a PLFC clause.